\documentclass[sigconf,natbib=true]{acmart}
\AtBeginDocument{%
  }

\copyrightyear{2026}
\acmYear{2026}
\setcopyright{cc}
\setcctype{by}
\acmConference[SIGIR '26]{How often do Answers Change? Estimating Recency Requirements in Question Answering}{July 20--24, 2026}{Melbourne, Australia}
\acmPrice{}
\acmDOI{XXXXXXX.XXXXXXX}
\acmISBN{978-1-4503-XXXX-X/2018/06}

\usepackage{booktabs}
\usepackage{multirow}
\usepackage{pifont}
\usepackage{xspace}
\usepackage{enumitem}
\usepackage[most]{tcolorbox}
\usepackage{tabularx}
\usepackage{adjustbox}
\usepackage{colortbl}
\usepackage{makecell}
\usepackage{graphicx}

\definecolor{customblue}{rgb}{0.168, 0.364, 0.557}
\newcommand{\RecencyQA}{\texttt{RECENCYQA}\xspace}

\usepackage[table,dvipsnames]{xcolor}

\newcommand{\cmark}{\ding{51}}
\newcommand{\xmark}{\ding{55}}

\newcommand{\pos}[1]{\cellcolor{green!10}\textcolor{ForestGreen}{\textbf{#1$\uparrow$}}}
\newcommand{\negp}[1]{\cellcolor{red!10}\textcolor{BrickRed}{\textbf{#1$\downarrow$}}}
\newcommand{\bp}[1]{\cellcolor{gray!20}\textcolor{black}{#1}}

\setlist[itemize]{leftmargin=*}

\begin{document}

\title{How often do Answers Change? Estimating Recency Requirements in Question Answering}

\author{Bhawna Piryani}
\orcid{0009-0005-3578-2393}
\affiliation{%
  \institution{University of Innsbruck}
  \city{Innsbruck}
  \country{Austria}
  }
\email{bhawna.piryani@uibk.ac.at}

\author{Zehra Mert}
\orcid{0009-0002-1038-985X}
\affiliation{%
  \institution{MEF University}
  \city{Istanbul}
  \country{Turkey}
  }
\email{mertz@mef.edu.tr}

\author{Adam Jatowt}
\orcid{0000-0001-7235-0665}
\affiliation{%
  \institution{University of Innsbruck}
  \city{Innsbruck}
  \country{Austria}
  }
\email{adam.jatowt@uibk.ac.at}


\begin{abstract}
  Large language models (LLMs) often rely on outdated knowledge when answering time-sensitive questions, leading to confident yet incorrect responses. Without explicit signals indicating whether up-to-date information is required, models struggle to decide when to retrieve external evidence, how to reason about stale facts, and how to rank answers by their validity. Existing benchmarks either periodically refresh answers or rely on fixed templates, but they do not reflect on how frequently answers change or whether a question inherently requires up-to-date information. To address this gap, we introduce a recency–stationarity taxonomy that categorizes questions by how often their answers change and whether this change frequency is time-invariant or context-dependent. Building on this taxonomy, we present \RecencyQA\footnote{\url{https://github.com/DataScienceUIBK/RecencyQA}}, a dataset of 4,031 open-domain questions annotated with recency and stationarity labels. Through human evaluation and empirical analysis, we show that non-stationary questions, i.e. those where context changes the recency requirement,
  are significantly more challenging for LLMs, with difficulty increasing as update frequency rises. By explicitly modeling recency and context dependence, \RecencyQA enables fine-grained benchmarking and analysis of temporal reasoning beyond binary notions of freshness, and provides a foundation for developing recency-aware and context-sensitive question answering systems.
\end{abstract}


\maketitle

\section{Introduction}
\label{sec:introduction}

Large language models (LLMs) have shown strong performance across a wide range of Question Answering (QA) tasks \cite{brown2020language,chang2024survey}. However, their ability to handle \emph{temporal questions}—questions whose answers change over time—remains limited. Most existing QA benchmarks treat answers as static facts \cite{joshi-etal-2017-triviaqa,kwiatkowski-etal-2019-natural}, overlooking the reality that many real-world information needs are inherently dynamic. For instance, \emph{What is the interest rate set by the European Central Bank?} requires a temporally grounded answer that reflects the current state of the world, rather than snapshots fixed at a model’s training cutoff.

Recent work in temporal QA has made progress but still leaves important gaps. Benchmarks such as TimeQA \cite{chen2021dataset}, TEMPLAMA \cite{dhingra-etal-2022-time}, and TIQ \cite{jia2024tiq} introduce time-sensitive questions, while others like RealTimeQA \cite{kasai2023realtime}, FreshQA \cite{vu-etal-2024-freshllms}, and PATQA \cite{meem-etal-2024-pat} periodically refresh answers to improve temporal validity. However, these datasets largely reduce temporal validity to a binary distinction—current versus outdated—without modeling how frequently answers are expected to change. As a result, they fail to distinguish between facts that update hourly, annually, or effectively never. Meanwhile, datasets such as TempReason \cite{tan-etal-2023-towards} and CRONQUESTIONS \cite{chen-etal-2023-multi} emphasize event-based temporal reasoning but do not model how quickly information becomes obsolete. As a result, existing benchmarks lack a principled way to estimate a question’s \emph{recency demand} or distinguish between stable and context-dependent temporal behavior.

To address these limitations, we introduce a \emph{recency–stationarity} taxonomy for characterizing temporal sensitivity in questions. This taxonomy describes questions along two orthogonal dimensions: (i) how often a question’s answer should be updated—ranging from rapidly changing facts to stable or permanent ones—and (ii) whether this expected update frequency is context-invariant (\emph{stationary}) or context-dependent (\emph{non-stationary}). This distinction captures not only how quickly information evolves, but also whether the “rate of change” in a question’s temporal relevance remains stable over time. For example, \emph{Who is the CEO of Twitter?} is typically low-frequency and stationary, whereas \emph{What is the inflation rate in the United States?} may become high-frequency and non-stationary during periods of economic instability.

\begin{table*}[!]
\small
\centering
\caption{Comparison of temporal question answering datasets. Abbreviations: Man.=created manually, Gen.=Automatically generated, Man.-Filt.=filtered from other datasets, Man.+Gen.=created by crowdsourcing and LLM generation, Templ.=created using templates, Man.-Filt.+Gen=filtered from other datasets and LLM generation, KC=Knowledge Corpus.}
\label{tab:datasetcomparison}
\resizebox{0.8\textwidth}{!}{
\begin{tabular}{l|c|c|c|c|c}
\toprule
\textbf{Dataset} & \textbf{Creation} & \textbf{KC} & \textbf{Multi-hop} & \textbf{Recency-Label} & \textbf{\# Ques.} \\
\midrule
TimeQA \citep{chen2021dataset} & Templ.-Wikidata & Wikipedia & \xmark & \xmark & 20k \\
SituatedQA \citep{zhang-choi-2021-situatedqa} & Man.-Filt. & Wikipedia & \xmark & \xmark & 12k \\
TempLama \citep{dhingra-etal-2022-time} & Templ./Cloze & Custom-News & \xmark &\xmark  & 50k \\
StreamingQA \citep{liska2022streamingqa} & Man.+Gen. & WMT News & \cmark & \xmark & 410k \\
ArchivalQA \citep{wang2022archivalqa} & Gen. & NYT Articles & \xmark & \xmark & 532k \\
ChroniclingAmericaQA \citep{piryani2024chroniclingamericaqa} & Gen. & Chronicling America Newpaers & \xmark & \xmark & 485k \\
RealTimeQA \citep{kasai2023realtime} & News websites & News Articles & \cmark & \xmark  & $\sim$5k \\
PATQA \citep{meem-etal-2024-pat} & Templ.-wikidata & Wikipedia & \cmark & \xmark& 6,172 \\
FreshQA \citep{vu-etal-2024-freshllms} & Man. & Google Search & \cmark & \xmark& 600 \\

\midrule
\textbf{RecencyQA (ours)} & Man.-Filt.+Gen & Wikipedia/Wikidata & \cmark &  \cmark & 4,031 \\
\bottomrule
\end{tabular}
}
\end{table*}



Based on this taxonomy, we present \textsc{\RecencyQA}—the first dataset featuring open-domain questions annotated with both recency and stationarity labels. Each question is paired with contextual information that enables temporal interpretation, allowing for both static and dynamic temporal evaluation. The dataset supports fine-grained analysis of temporal behavior in LLMs across a broad spectrum of recency demands. In our experiments, we use the \textsc{RecencyQA} dataset to evaluate how well state-of-the-art LLMs reason about temporal dynamics, both with and without explicit contextual grounding.

In summary, we make the following contributions.

\begin{itemize}
    \item We introduce a two-dimensional taxonomy that characterizes questions by (i) the expected update frequency of their answers and (ii) whether this temporal behavior is context-invariant (\emph{stationary}) or context-dependent (\emph{non-stationary}).

    \item We release \RecencyQA, the first dataset of questions annotated with recency distributions, stationarity labels, and temporal context paired with different recency labels, enabling context-dependent evaluation.

    \item We perform comprehensive analyses and experiments to validate the quality of \RecencyQA and demonstrate its effectiveness for evaluating temporal awareness and recency reasoning in question answering systems.
\end{itemize}

\section{Related Work}
\label{sec:related_work}

Early QA benchmarks such as TriviaQA \cite{joshi-etal-2017-triviaqa}, Natural Questions \cite{kwiatkowski-etal-2019-natural}, or WebQ \cite{berant-etal-2013-semantic} assume answers to be static, overlooking that many real-world answers evolve over time. Temporal QA datasets such as TimeQA \cite{chen2021dataset}, ArchivalQA \cite{wang2022archivalqa}, ChroniclingAmericaQA \cite{piryani2024chroniclingamericaqa}, and SituatedQA \cite{zhang-choi-2021-situatedqa} introduced timestamped or context-anchored questions to assess models’ temporal reasoning. Later efforts, including StreamingQA \cite{liska2022streamingqa}, RealTimeQA \cite{kasai2023realtime}, PATQA \cite{meem-etal-2024-pat}, and FreshQA \cite{vu-etal-2024-freshllms}, emphasize answer freshness by continually updating data or sourcing recent information. However, these benchmarks typically frame recency as a binary distinction between current and outdated information, without modeling the expected rate at which answers change.

Parallel work on temporal validity and factual drift studies how information and model behavior evolve over time. \citet{zhao-etal-2022-impact} examines explanation degradation under evolving data, while \citet{wenzel-jatowt-2024-temporal} predict when factual statements become outdated. \citet{dhingra-etal-2022-time} model language models as temporal knowledge bases, and \citet{yang-etal-2020-improving} predict fact duration. These approaches focus on fact-level temporal dynamics and do not characterize question-level information needs by their expected update frequency or contextual variability, which our taxonomy explicitly models.

In the context of Temporal Information Retrieval (TIR) the Temporalia benchmark series \cite{DBLP:conf/ntcir/JohoJBNY14, DBLP:conf/ntcir/JohoJBYY16} and related work \cite{li2003time, berberich2010language, styskin2011recency} classify queries into coarse temporal categories (e.g., Past, Recency, Future, Atemporal) and apply recency-sensitive ranking for time-critical information needs. Similarly, retrieval-augmented generation (RAG) methods \cite{lewis2020retrieval, izacard-grave-2021-leveraging} mitigate factual staleness by incorporating external evidence, but lack mechanisms to reason about when retrieval is necessary based on a question’s recency demand.

Across QA, retrieval, and drift prediction, existing approaches treat temporal sensitivity as discrete, static, or reactive. In contrast, we propose a Recency–Stationarity Taxonomy that models both how frequently answers are expected to change and whether that rate is stable or context-dependent. This enables a proactive, fine-grained analysis of temporal reliability in LLMs.

While prior datasets focus on when information is valid, they do not capture how often it changes or whether that rate is context-sensitive. \RecencyQA complements these efforts by introducing recency and stationarity annotations that go beyond binary freshness. Table~\ref{tab:datasetcomparison} summarizes key differences between \RecencyQA and existing temporal QA benchmarks.


\section{Taxonomy}
\label{sec:taxonomy}

Temporal dynamics play a crucial role in question answering (QA), as the validity of an answer often depends on when the question is asked. Yet most existing QA benchmarks still assume that answers are static and permanently correct, overlooking that many real-world questions are inherently time-dependent.

To address this, we introduce a \textbf{Recency–Stationarity Taxonomy} to characterize the temporal sensitivity of questions along two orthogonal dimensions: \textit{recency} and \textit{stationarity}.
\textbf{Recency} captures the expected temporal stability of a question’s answer—that is, how soon the answer is likely to change under typical real-world conditions. For example, for \textit{Who is the president of the United States?}, the answer typically changes every few years, while \textit{What is the chemical symbol for gold?} remains constant indefinitely.

Complementing recency is the concept of \textbf{Stationarity}, which refers to the stability of a question's recency behavior over time. A question is \emph{stationary} if its recency label remains consistent—for instance, \textit{When is the World Technology Summit held?} always falls into an annual update cycle. In contrast, a \emph{non-stationary} question, such as \textit{Who is leading the Olympic medal table?}, exhibits different recency patterns depending on the temporal context—high-frequency updates during the Olympics, and none between events. Importantly, stationarity does not measure how often an answer changes, but whether that frequency itself remains consistent across time.

In summary, the recency label measures how often an answer is expected to change, while stationarity captures whether that update frequency is consistent or varies over time.

To operationalize this taxonomy, we define twelve discrete recency classes representing the expected time until a question’s answer changes, as shown in Table~\ref{tab:recency_taxonomy}. Stationarity, in turn, determines whether a question consistently belongs to the same recency class or shifts based on context.

This taxonomy provides a principled and operational framework for quantifying answer volatility in QA, enabling systematic analysis of temporal reliability and recency awareness in large language models.

\begin{table}[!]
\centering
\small
\caption{
The proposed Recency Taxonomy consists of twelve classes,
ordered from highly volatile (top) to temporally stable (bottom).
Each class reflects the expected time until a question’s answer first changes.
}
\begin{tabular}{lc}
\toprule
\textbf{Recency Class} & \textbf{Expected Time Until Answer Change} \\
\midrule
An-Hour        & Within an hour \\
A-Few-Hours    & Within a few hours \\
A-Day          & Within a day \\
A-Few-Days     & Within a few days \\
A-Week         & Within a week \\
A-Few-Weeks    & Within a few weeks \\
A-Month        & Within a month \\
A-Few-Months   & Within a few months \\
A-Year         & Within a year \\
A-Few-Years    & Within a few years \\
Many-Years     & After many years \\
Never          & Not expected to change \\
\bottomrule
\end{tabular}

\label{tab:recency_taxonomy}
\end{table}

\section{RecencyQA Dataset}
\label{sec:dataset_construction}
Based on the taxonomy introduced in Section~\ref{sec:taxonomy}, we construct the \textsc{RecencyQA} dataset to enable fine-grained evaluation of temporal sensitivity in question answering. Unlike prior benchmarks that treat temporal validity as binary, \textsc{RecencyQA} explicitly models both (i) the expected time until an answer changes (recency) and (ii) whether this update behavior remains stable across contexts (stationarity). The construction pipeline consists of three main stages: (1) question sampling, (2) recency and stationarity label generation, and (3) Temporal Context Generation. We describe the construction pipeline in detail below. 

We adopt \textsc{LLaMA 3.3 (70B)}~\cite{grattafiori2024llama} as the primary model for generation throughout the pipeline. This model demonstrates strong performance across a range of NLP tasks and is fully open-source and accessible via Hugging Face\footnote{\url{https://huggingface.co/meta-llama/Llama-3.3-70B-Instruct}}, ensuring reproducibility and facilitating adoption of our pipeline for future datasets. 

\subsection{Question Selection}

To construct a diverse and temporally grounded question set, we sample questions from two complementary sources:
(i) existing temporal QA datasets and 
(ii) event-based questions generated by a large language model.

For the first source, we use three widely adopted temporal QA datasets: FreshQA~\cite{vu-etal-2024-freshllms}, PATQA~\cite{meem-etal-2024-pat}, and SituatedQA~\cite{zhang-choi-2021-situatedqa}. These were selected because they explicitly emphasize temporal reasoning and collectively cover a broad range of answer update behaviors. We sample a total of 2,988 questions using dataset-specific criteria. From FreshQA, we include all questions labeled with a \emph{true premise} (453 questions). From PATQA, we select 40 questions per relation type, yielding 1,280 questions. From SituatedQA, which focuses on context-sensitive temporal reasoning, we incorporate the entire test split (931 questions).

To expand coverage beyond existing benchmarks, we generate additional questions using \textsc{LLaMA 3.3 (70B)}. We first prompt the model to produce around 120 brief descriptions of real-world events. Each description was then used as a seed for generating 20 time-sensitive questions, resulting in approximately 2,400 additional questions grounded in real-world scenarios. After generation, we removed 173 duplicate questions. The prompt for event and question generation is given in our GitHub repository. Combining both sources yields a final collection of 4,891 unique questions.

\subsection{Recency and Stationarity Label Generation}
\label{sec:multi_sample_recency}

In this module, we assign a recency label to each question, capturing the expected time until its answer changes according to the 12-class taxonomy defined in Section~\ref{sec:taxonomy}. To estimate recency more robustly, we prompt \textsc{LLaMA 3.3 (70B)} 13 times per question and aggregate the outputs to obtain a distribution over recency classes.

We then determine a stationarity label for each question, indicating whether its recency classification remains stable over time or varies depending on the temporal context. The details of this procedure are described in the following subsection.

Finally, conditioned on the assigned recency and stationarity labels, we generate a corresponding temporal context that concretely grounds the question’s temporal behavior. Each component of this process is described below.

\subsubsection{Recency Label Generation.}
\label{sec:recency_label_generation}
For each question $q$, we prompt \textsc{LLaMA 3.3 (70B)} 13 times to assign a recency class from the predefined 12-category taxonomy shown in Table~\ref{tab:recency_taxonomy}. Each prompt is sampled independently, yielding multiple recency annotations per question. The taxonomy spans a broad spectrum of temporal dynamics, from rapidly changing information (e.g., hourly updates) to effectively permanent facts.

We use 13 samples because the taxonomy contains 12 discrete classes. Sampling one more time than the number of classes ensures that a strict majority label can always be identified, even in cases of near-uniform disagreement across classes.

Rather than producing a single deterministic label, this procedure yields an empirical distribution over recency classes, capturing uncertainty and variation in temporal interpretation. We assign the majority class as the primary recency label and retain the full distribution for subsequent analysis, particularly for identifying non-stationary questions. At this stage, each question is associated with a primary recency label and its corresponding recency distribution.

\begin{table}
\small
\caption{Statistical Summary of the \RecencyQA Dataset}
\label{tbl:dataset_statistics}
\resizebox{0.9\columnwidth}{!}{%
\scriptsize
\begin{tabular}{@{}l@{\hspace{90pt}}c@{}}
\toprule
Metric                                              & Value \\ 
\midrule
Total Recency classes                                 &  12  \\
Total Questions                                     &  4,031   \\
Total Recency Labels                                &  5,237  \\ 
\midrule
Total Stationary Questions                          & 2,910      \\
Total Non-Stationary Questions                   &  1,121    \\ 
\midrule
Total Singlehop Questions                            & 3,161\\
Total Multihop Questions                           & 8,70\\ 
\midrule
Avg. Question Length (words)                        &  14.26   \\
Avg. Context Length (words)                   &   22.22   \\ 

\bottomrule
\end{tabular}%
}
\end{table}

\subsubsection{Stationary Label.}
\label{sec:stationarity_label}

To assess the temporal stability of each question, we assign a stationarity label indicating whether the expected rate of answer change remains consistent over time (stationary) or varies depending on temporal context (non-stationary). A question is considered stationary if its recency classification is stable across time, and non-stationary if the classification varies with context.

To obtain an initial stationarity judgment, we prompt three state-of-the-art LLMs-GPT 5.2 \cite{singh2025openai}, Gemini 3 Flash \cite{team2023gemini}, and Claude Sonnet 4.5—to independently classify each question as stationary or non-stationary. The majority vote among the three models serves as the preliminary label for stationarity.

We then cross-validate this preliminary label against the empirical recency distribution derived in Section~\ref{sec:recency_label_generation}. We applied the following consistency rules: (1) If a question receives a single recency class across all 13 samples and the preliminary label is stationary, we retain the stationary label. (2) If multiple recency classes are observed and the preliminary label is non-stationary, we retain the non-stationary label. (3) If multiple recency classes are observed but the preliminary label is stationary, we override it and reclassify the question as non-stationary, since label variation indicates temporal instability. (4) If a single recency class is observed but the preliminary label is non-stationary, we discard the question due to the inconsistency between recency dynamics and stationarity judgment.

This cross-validation procedure enforces alignment between recency variability and stationarity classification, improving annotation reliability. After this filtering step, 4,031 questions remain with valid recency and stationarity labels.

\subsubsection{Temporal Context Generation}

To ground each question’s temporal behavior in realistic scenarios, we generate short temporal contexts designed to induce specific recency interpretations. For each question, we generate three independent contexts corresponding to each validated recency label. Stationary questions are associated with a single recency label, whereas non-stationary questions may have multiple labels, each requiring a distinct context. In total, this process produces approximately 15,711 candidate contexts.

To ensure that the generated contexts reliably induce the intended temporal interpretation, we verify each question–context pair by repeatedly relabeling. Specifically, we prompt LLaMa 3.3 (70B) 13 times with the question–context pair and request a recency classification. A context is accepted only if the intended recency label is reproduced in all of the 13 runs (i.e., a strict majority). If the dominant label differs from the intended one, the context is regenerated. This procedure is repeated until every recency label associated with a question has at least one verified context.

Finally, to reduce redundancy, we retain a single verified context per question. When multiple valid contexts are available, we select the longest one, as longer contexts typically provide richer temporal grounding. This verification strategy prioritizes precision over recall, minimizing the retention of contexts with ambiguous or misleading temporal cues.

We compile all retained questions along with their associated metadata, including the 13 recency label samples, the majority recency label, the full recency distribution, the stationarity label, and the verified temporal context. These fields are stored in a structured JSON format, constituting the final release of the \RecencyQA dataset.


\section{Dataset Analysis}

In this section, we analyze the \RecencyQA dataset from two perspectives: \textit{Dataset statistics} and \textit{Human Evaluation}.

\subsection{Dataset Statistics}
\label{sec:dataset_statistics}

\begin{table}[!]
\centering
\small
\caption{Human evaluation results for the \RecencyQA dataset. 
The upper section reports labeling accuracies, and the lower section reports mean human ratings (1-5 scale). 
Recency accuracy is shown as strict and tolerant (within $\pm$1 bin).}
\label{tbl:human_results}
\resizebox{0.8\columnwidth}{!}{%
\begin{tabular}{lcc}
\toprule
 & \textbf{Accuracy} & \textbf{Tolerant Accuracy} \\ 
\midrule
Recency &0.76  &0.81  \\
Stationarity &0.78  & -- \\ 
\midrule
\multicolumn{3}{c}{\textbf{Human Rating Averages (1-5 scale)}} \\ 
\midrule
Clarity of Question & Difficulty to Label & Difficulty to Answer \\
 4.66 &2.26 & 2.41\\
\bottomrule
\end{tabular}%
}
\end{table}

After applying all the verification and filtering steps we obtain \RecencyQA dataset comprising 4,031 questions. Each question is annotated with a stationarity label, one or more recency labels, and a verified temporal context that grounds the assigned recency interpretation. Table~\ref{tbl:dataset_statistics} presents a detailed statistical summary of the dataset across key features. 

\subsection{Human Evaluation}

We conducted a human evaluation of the \RecencyQA dataset to assess whether the LLM-generated labels, given the verified temporal context, align with human judgments. Six graduate students (3 male, 3 female) participated in the study.

We sampled 240 questions, ensuring balanced coverage across recency and stationarity labels. The questions were divided into six sets of 40 questions, with each set independently annotated by three annotators. For each question, annotators were asked to: (i) assign a \textit{recency} label based on the provided context, (ii) assign a \textit{stationarity} label, (iii) rate the \textit{clarity} of the question, (iv) rate the \textit{difficulty of assigning a recency label}, and (v) rate the \textit{difficulty of answering} the question. Ratings were provided on a 5-point scale. For clarity, 1 indicated “not clear” and 5 indicated “very clear.” For difficulty ratings, 1 denoted “very easy” and 5 denoted “very difficult.” For label aggregation, we applied majority voting to obtain final human-assigned recency and stationarity labels. These majority labels were compared against the corresponding dataset annotations to compute agreement accuracy.

Table~\ref{tbl:human_results} summarizes the quantitative evaluation results.
The results indicate high agreement between dataset labels and human judgments for both recency and stationarity, supporting the reliability of the annotation pipeline. Clarity ratings are consistently high, while moderate difficulty scores suggest that the labeling task is understandable yet non-trivial.


\begin{table}[!]
\centering
\caption{
Performance comparison of different LLMs on recency classification across different prompting paradigms. Gray cells indicate the best results within each prompting strategy, while underlined values represent the overall best performance across all approaches for each metric. 
}
\label{tbl:temporal_classification}
\begin{adjustbox}{width=\columnwidth}
\begin{tabular}{lccccc}
\toprule
\textbf{Model} & \textbf{\# of Parameters} & \textbf{Accuracy} & \textbf{Tolerant Accuracy} & \textbf{F1 Score} & \textbf{Tolerant F1 Score} 
 \\
\midrule
\multicolumn{6}{c}{\textbf{Zero-shot Prompting}} \\
\midrule
Qwen 2.5 & 7B   &0.3001	&0.5975	&0.2514	&0.748
    \\
LLaMA 3 & 8B   &0.3467	&0.6145	&0.3381	&0.7612   \\
Mistral  & 24B   &0.2878	&0.6224	&0.2722	&0.7673   \\
Gemma 3 & 27B   &\bp{0.403}	&\bp{0.646}	&\bp{0.4224}	&\bp{0.7849}    \\
Apertus & 70B   &0.3032	&0.6249	&0.3139	&0.7692    \\
Qwen 2.5 & 72B   &0.3681 &0.761	&0.3302	&0.8643    \\
\midrule
\multicolumn{6}{c}{\textbf{Few-shot Prompting}} \\
\midrule
Qwen 2.5 & 7B   &0.3557	&0.5915	&0.3515	&0.7433     \\
LLaMA 3 & 8B   &0.0275	&0.0603	&0.0045	&0.1137
   \\
Mistral  & 24B   &0.1595	&0.3686	&0.1269	&0.5387
   \\
Gemma 3 & 27B   &\bp{\underline{0.5205}}	&\bp{\underline{0.7891}}	&\bp{\underline{0.5145}}	&\bp{\underline{0.8821}}
   \\
Apertus & 70B   &0.3793	&0.681	&0.3698	&0.8102    \\
Qwen 2.5 & 72B   &0.5048	&0.7809	&0.4764	&0.877
   \\
\midrule
\multicolumn{6}{c}{\textbf{Chain-of-Thought (CoT) Prompting}} \\
\midrule
Qwen 2.5 & 7B   &0.2417	&0.5053	&0.1972	&0.6714     \\
LLaMA 3 & 8B   &0.3659	&0.6828	&0.3412	&0.8115   \\
Mistral  & 24B   &0.3763	&\bp{0.7663}	&0.3296	&\bp{0.8677}   \\
Gemma 3 & 27B   &\bp{0.3878}	&0.6557	&\bp{0.4168}	&0.7921    \\
Apertus & 70B   &0.2166	&0.5375	&0.2461	&0.6992   \\
Qwen 2.5 & 72B   &0.4336	&0.8105	&0.4264	&0.8953
    \\

\bottomrule
\end{tabular}
\end{adjustbox}
\end{table}

\begin{table*}[!]
\centering
\caption{
Effect of context on stationary and non-stationary question performance across prompting strategies.
Arrows ($\uparrow$ / $\downarrow$) and colored percentages indicate improvement or degradation. 
}
\label{tab:context_stationarity_vs_nonstaionary}
\begin{adjustbox}{width=0.8\textwidth}
\begin{tabular}{lc|cccccc||ccccccc}
\toprule
\textbf{Model} & \textbf{\# of Parameters} &
\multicolumn{3}{c}{\textbf{Stationary (Accuracy)}} &
\multicolumn{3}{c}{\textbf{Stationary (F1-score)}} &
\multicolumn{3}{c}{\textbf{Non-stationary (Accuracy)}} &
\multicolumn{3}{c}{\textbf{Non-stationary (F1-score)}} \\
\cmidrule(lr){3-5} \cmidrule(lr){6-8} \cmidrule(lr){9-11} \cmidrule(lr){12-14}
 & & \textbf{w/ C} & \textbf{w/o C} & \textbf{$\Delta$\%} &
     \textbf{w/ C} & \textbf{w/o C} & \textbf{$\Delta$\%} &
     \textbf{w/ C} & \textbf{w/o C} & \textbf{$\Delta$\%} &
     \textbf{w/ C} & \textbf{w/o C} & \textbf{$\Delta$\%} \\
\midrule
\multicolumn{14}{c}{\textbf{Zero-shot Prompting}} \\
\midrule
Qwen2.5-7B & 7B & 0.309 & 0.318 & \negp{-2.8\%} & 0.240 & 0.256 & \negp{-6.1\%} & 0.251 & 0.255 & \negp{-1.5\%} & 0.178 & 0.181 & \negp{-1.8\%} \\
Llama-3.1-8B & 8B & 0.429 & 0.401 & \pos{+6.9\%} & 0.383 & 0.366 & \pos{+4.6\%} & 0.296 & 0.205 & \pos{+44.3\%} & 0.284 & 0.212 & \pos{+33.7\%} \\
Mistral-24B & 24B & 0.332 & 0.308 & \pos{+7.8\%} & 0.298 & 0.285 & \pos{+4.5\%} & 0.286 & 0.235 & \pos{+22.1\%} & 0.216 & 0.177 & \pos{+22.0\%} \\
gemma-3-27b & 27B & 0.435 & 0.456 & \negp{-4.5\%} & 0.430 & 0.443 & \negp{-3.0\%} & 0.285 & 0.267 & \pos{+6.9\%} & 0.256 & 0.230 & \pos{+11.7\%} \\
Apertus-70B & 70B & 0.290 & 0.315 & \negp{-7.8\%} & 0.292 & 0.319 & \negp{-8.4\%}& 0.256 & 0.273 & \negp{-6.2\%} & 0.221 & 0.217 & \pos{+1.6\%} \\
Qwen2.5-72B & 72B & 0.376 & 0.378 & \negp{-0.5\%} & 0.351 & 0.316 & \pos{+11.0\%} & 0.404 & 0.343 & \pos{+17.7\%} & 0.310 & 0.253 & \pos{+22.6\%} \\
\midrule
\multicolumn{14}{c}{\textbf{Few-shot Prompting}} \\
\midrule
Qwen2.5-7B & 7B & 0.458 & 0.395 & \pos{+16.0\%} & 0.407 & 0.363 & \pos{+12.1\%} & 0.343 & 0.266 & \pos{+29.1\%} & 0.290 & 0.209 & \pos{+39.0\%} \\
Llama-3.1-8B & 8B & 0.359 & 0.344 & \pos{+4.4\%} & 0.281 & 0.294 & \negp{-4.7\%} & 0.351 & 0.240 & \pos{+46.0\%} & 0.256 & 0.184 & \pos{+39.3\%} \\
Mistral-24B & 24B & 0.152 & 0.173 & \negp{-12.1\%} & 0.032 & 0.132 & \negp{-75.4\%} & 0.053 & 0.124 & \negp{-56.9\%} & 0.016 & 0.089 & \negp{-82.0\%} \\
gemma-3-27b & 27B & 0.564 & 0.561 & \pos{+0.6\%} & 0.511 & 0.537 & \negp{-4.7\%} & 0.494 & 0.415 & \pos{+19.1\%} & 0.420 & 0.349 & \pos{+20.1\%} \\

Apertus-70B & 70B & 0.284 & 0.210 & \pos{+35.6\%} & 0.303 & 0.209 & \pos{+45.0\%} & 0.258 & 0.245 & \pos{+5.4\%} & 0.321 & 0.264 & \pos{+21.3\%} \\

Qwen2.5-72B & 72B & 0.619 & 0.582 & +6.4\% & 0.549 & 0.486 & +13.0\% & 0.418 & 0.345 & +21.2\% & 0.396 & 0.284 & +39.3\% \\
\midrule
\multicolumn{14}{c}{\textbf{Chain-of-Thought (CoT) Prompting}} \\
\midrule

Qwen2.5-7B & 7B & 0.270 & 0.249 & \pos{+8.6\%} & 0.197 & 0.198 & \negp{-0.7\%} & 0.230 & 0.224 & \pos{+2.8\%} & 0.136 & 0.153 & \negp{-10.9\% }\\
Llama-3.1-8B & 8B & 0.414 & 0.384 & \pos{+7.8\%} & 0.377 & 0.348 & \pos{+8.5\%} & 0.385 & 0.319 & \pos{+20.7\%} & 0.295 & 0.243 & \pos{+21.3\%} \\
Mistral-24B & 24B & 0.392 & 0.389 & \pos{+0.8\%} & 0.344 & 0.339 & \pos{+1.5\%} & 0.359 & 0.343 & \pos{+4.9\%} & 0.254 & 0.225 & \pos{+12.8\%} \\
gemma-3-27b & 27B & 0.436 & 0.426 & \pos{+2.5\%} & 0.438 & 0.437 & \pos{+0.1\%} & 0.323 & 0.290 & \pos{+11.4\%} & 0.266 & 0.226 & \pos{+17.4\%} \\
Apertus-70B & 70B & 0.214 & 0.204 & \pos{+4.7\%} & 0.253 & 0.245 & \pos{+3.5\%} & 0.242 & 0.248 & \negp{-2.5\%} & 0.209 & 0.166 & \pos{+25.7\%} \\
Qwen2.5-72B & 72B & 0.465 & 0.461 & \pos{+0.9\%} & 0.443 & 0.431 & \pos{+2.9\%} & 0.426 & 0.363 & \pos{+17.4\%} & 0.302 & 0.249 & \pos{+21.1\%} \\
\bottomrule
\end{tabular}
\end{adjustbox}
\end{table*}

\section{Experimental Setup}
\label{sec:experimental_setup}

We evaluate six general-purpose large language models (LLMs) spanning diverse architectures and parameter scales to reduce model-family or size bias. The evaluated models include Qwen-2.5 (7B and 72B) \cite{yang2025qwen3}, LLaMA-3 (8B) \cite{grattafiori2024llama}, Mistral (24B) \cite{jiang2023mistral}, Gemma 3 (27B) \cite{team2025gemma}, and Apertus (70B) \cite{swissai2025apertus}.

For evaluation, we use two standard metrics: \textbf{Accuracy} and \textbf{F1 Score}. Additionally, we report \textbf{Tolerant Accuracy} and \textbf{Tolerant F1}, which treat predictions as partially correct if the predicted recency class is adjacent to the ground-truth label. As recency classes follow a natural temporal progression (e.g., from “an hour” to “a few hours” to “never”), the tolerant evaluation captures a more nuanced and realistic measure of performance, accounting for the gradual and overlapping nature of temporal boundaries.

All experiments were conducted using Hugging Face model implementations, with a decoding temperature of 0.5 to balance stability and reasoning diversity. We evaluated three prompting strategies: zero-shot, few-shot, and chain-of-thought. The prompts used in experiments are available in our public GitHub repository.

\section{Experiments and Results}
\label{sec:experimental_results}

We evaluate the effectiveness of \textsc{RecencyQA} as a benchmark for measuring temporal awareness in large language models. Specifically, we assess model performance across three complementary tasks: (1) temporal classification, (2) context sensitivity, and (3) dynamic recency transition (RL$_1 \rightarrow$ RL$_2$).







\paragraph{\textbf{Recency Classification}}
\label{sec:temporal_classification}

We first examine whether models can infer recency properties directly from the question without additional context. This setting evaluates models’ intrinsic ability to estimate the expected update frequency of an answer.

Given a question, the model predicts one of the 12 predefined recency classes from our taxonomy. No temporal context is provided; only the question text is used as input. Table~\ref{tbl:temporal_classification} reports performance across models and prompting strategies.

\paragraph{Analysis.}
Recency classification proves to be a challenging task across all models.  With accuracy ranging from 24\% to 52\%. Few-shot prompting yields the strongest performance, with Gemma 3 (27B) achieving the highest accuracy (52.05\%) and macro F1 (51.45\%). This demonstrates that providing a small number of examples helps models better align with the notion of answer update frequency (recency frequency). Tolerant metrics are substantially higher than strict metrics (best Tolerant F1 > 0.88), indicating that models often predict temporally adjacent classes rather than completely incorrect ones. 
\textit{Overall, while LLMs capture coarse temporal ordering, fine-grained recency classification remains difficult.}

\paragraph{\textbf{Role of Context}}
\label{sec:role_of_context}

We next investigate how explicit temporal context influences recency prediction performance. For each question, we evaluate two settings: \textbf{Without Context}, where only the question is provided, and \textbf{With Context}, where the verified temporal context is appended. Results are analyzed separately for stationary and non-stationary questions to assess whether context differentially impacts stable versus context-dependent temporal behavior. Table~\ref{tab:context_stationarity_vs_nonstaionary} reports performance under both settings and the corresponding performance differences.

\paragraph{Analysis.}
The effect of context differs substantially between stationary and non-stationary questions. For non-stationary questions, incorporating temporal context generally improves performance across models and prompting strategies, with gains exceeding +40\% in certain cases. This suggests that models can leverage contextual cues when recency expectations are context-dependent. By contrast, for stationary questions, performance often remains unchanged or decreases when context is introduced. This pattern indicates that models may assign undue weight to temporal cues even when the underlying recency label is stable. In several few-shot configurations, performance degradation is particularly pronounced, pointing to interactions between in-context examples and appended temporal information. Taken together, \textit{these results show that contextual grounding enhances performance when temporal variability is intrinsic to the question, but may negatively affect prediction when temporal dynamics are stable.}

\begin{table}[!]
\centering
\small
\caption{
Performance on the recency transition task (RL$_1 \rightarrow$ RL$_2$) for Gemma 3 (27B). RL$_1$ and RL$_2$ report per-context accuracy, while Transition Accuracy measures successful adaptation across contexts.
}

\label{tbl:transition_results}
\begin{adjustbox}{width=\columnwidth}
\begin{tabular}{lccc}
\toprule
\textbf{Strategy} & \textbf{RL1 Accuracy} & \textbf{RL2 Accuracy} & \textbf{Transition Accuracy} \\ 
\midrule
Zero-shot &0.3348&0.205	&0.034
 \\
Few-Shot &0.4665&0.4505	&0.1481
  \\ 
Chain-of-Thought (COT) &0.3688	&0.3491	&0.0911
  \\ 
\bottomrule
\end{tabular}
\end{adjustbox}
\end{table}

\paragraph{\textbf{Recency Transition (RL$_1 \rightarrow$ RL$_2$)}}
\label{sec:recency_transition}

We evaluate models’ ability to adapt to dynamic temporal shifts through a recency transition task (RL$_1 \rightarrow$ RL$_2$), conducted exclusively on non-stationary questions. These questions are associated with multiple validated recency labels in \textsc{RecencyQA} dataset. For each question, we construct two contextual variants: C$_1$ corresponding to recency label RL$_1$, and C$_2$ corresponding to a different recency label RL$_2$. The correct label, therefore, differs across contexts, requiring the model to revise its prediction. We report accuracy for C$_1$ and C$_2$ separately and introduce \textit{transition accuracy}, defined as the percentage of cases in which a model correctly predicts RL$_1$ under C$_1$ and RL$_2$ under C$_2$.

\paragraph{Analysis.} Table~\ref{tbl:transition_results} presents results for Gemma 3 (27B), the best-performing model in the recency classification task. Although Gemma achieves moderate accuracy within individual contexts, transition accuracy remains substantially lower across all prompting strategies. For instance, under few-shot prompting, per-context accuracy exceeds 45\%, while transition accuracy reaches only 14.8\%. This pronounced gap indicates limited ability to update predictions in response to contextual change. \textit{The results suggest that, despite reasonable per-context performance, current LLMs exhibit predominantly static temporal reasoning and struggle with dynamic recency adaptation.}




\paragraph{Overall Discussion.}
Our results show that \textsc{RecencyQA} reveals systematic weaknesses in current LLMs’ temporal reasoning. Models struggle with fine-grained recency classification, respond inconsistently to contextual information, and show limited ability to adapt recency transitions. These findings indicate that temporal sensitivity is not reliably captured by existing systems.

By explicitly modeling both recency (expected update frequency) and stationarity (context dependence), \textsc{RecencyQA} enables controlled evaluation of temporal awareness at three levels: recency requirement classification, context sensitivity, and dynamic adaptation. The dataset, therefore, provides a structured benchmark for analyzing when temporal information should influence predictions and when it should not. Beyond benchmarking, \textsc{RecencyQA} supports research on recency-aware retrieval, temporal confidence calibration, freshness-sensitive ranking, and dynamic multi-hop temporal reasoning. It establishes a foundation for developing and evaluating temporally adaptive QA and retrieval systems.

\section{Conclusion}

We introduced \textsc{RecencyQA}, a benchmark and taxonomy for modeling how often answers change and whether their temporal behavior remains stable across contexts. By jointly capturing recency and stationarity, the dataset enables systematic evaluation of temporal sensitivity in question answering systems. Our analyses show that while LLMs perform reasonably well on temporally stable questions, their performance degrades as recency demands increase. Context improves results in dynamic settings but often harms performance on stationary questions, revealing a lack of temporal selectivity. \textsc{RecencyQA} provides a foundation for recency-aware QA and retrieval systems, encouraging future models that can determine when to rely on stored knowledge and when to seek updated information.


\bibliographystyle{ACM-Reference-Format}
\balance
\bibliography{sample-base}


\end{document}